\documentclass[runningheads,dvipsnames]{ECML2023_files/llncs}
\usepackage{amsmath,amsfonts,bm}

\def\eqref#1{equation~\ref{#1}}
\def\1{\bm{1}}

\DeclareMathAlphabet{\mathsfit}{\encodingdefault}{\sfdefault}{m}{sl}
\SetMathAlphabet{\mathsfit}{bold}{\encodingdefault}{\sfdefault}{bx}{n}

\DeclareMathOperator*{\argmin}{arg\,min}

\usepackage{hyperref}
\usepackage{url}

\newcommand{\norm}[1]{\left\|#1\right\|}

\usepackage[vlined,ruled]{algorithm2e}

\usepackage[utf8]{inputenc} % allow utf-8 input
\usepackage[T1]{fontenc}    % use 8-bit T1 fonts
\usepackage{hyperref}       % hyperlinks
\usepackage{url}            % simple URL typesetting
\usepackage{booktabs}       % professional-quality tables
\usepackage{amsfonts}       % blackboard math symbols
\usepackage{nicefrac}       % compact symbols for 1/2, etc.
\usepackage{microtype}      % microtypography
\usepackage{thm-restate}

\usepackage{microtype}
\usepackage{graphicx}
\usepackage{subfigure}
\usepackage{changes}

\usepackage{amsmath}
\usepackage{multirow}
\usepackage{hhline}
\usepackage{wrapfig}
\usepackage{floatrow}
\usepackage{caption}
\usepackage{enumitem}
\usepackage[export]{adjustbox}
\usepackage{amsmath,mathtools}

\usepackage{algorithmic}

\makeatletter
\newcommand{\raisemath}[1]{\mathpalette{\raisem@th{#1}}}
\newcommand{\raisem@th}[3]{\raisebox{#1}{$#2#3$}}
\makeatother

\DeclarePairedDelimiter\abs{\lvert}{\rvert}%

\newcommand{\uglad}{{\texttt{uGLAD}~}}
\newcommand{\ugladns}{{\texttt{uGLAD}}}

\newcommand{\ngm}{{\texttt{NGM~}}}
\newcommand{\fngm}{{\texttt{FedNGM~}}}
\newcommand{\fngms}{{\texttt{FedNGMs~}}}
\newcommand{\fngmns}{{\texttt{FedNGM}}}
\newcommand{\ngms}{{\texttt{NGMs~}}}
\newcommand{\ngmns}{{\texttt{NGM}}}
\newcommand{\ngmsns}{{\texttt{NGMs}}}
\newcommand{\ngr}{{\texttt{NGR~}}}
\newcommand{\ngrs}{{\texttt{NGRs~}}}

\newcommand{\ngrsns}{{\texttt{NGRs}}}

\author{%\hspace{-5mm}
  Urszula Chajewska~~~~%\Annd
  Harsh Shrivastava%\And
 }

 \institute{Microsoft Research, Redmond, USA}
\titlerunning{Federated Learning with Neural Graphical Models}
\begin{document}

%\twocolumn[
% \title{Federated Learning with Neural Graphical Models and Neural Graph Revealers}
\title{Federated Learning with Neural Graphical Models}

\maketitle

\begin{abstract}
Federated Learning (FL) addresses the need to create models based on proprietary data in such a way that multiple clients retain exclusive control over their data, while all benefit from improved model accuracy due to pooled resources. Recently proposed Neural Graphical Models (\ngmsns) and Neural Graph Revealers (\ngrsns) are Probabilistic Graphical Models that utilize the expressive power of neural networks to learn complex non-linear dependencies between the input features. They learn to capture the underlying data distribution and have efficient algorithms for inference and sampling. We develop a FL framework which maintains a global \ngmns/\ngr model that learns the averaged information from the local \ngmns/\ngr models while the training data is kept within the client's environment. Our design, \fngmns, avoids the pitfalls and shortcomings of neuron matching frameworks like Federated Matched Averaging that suffers from model parameter explosion. Our global model size doesn't grow with the number or diversity of clients. In the cases where clients have local variables that are not part of the combined global distribution, we propose a \textit{Stitching} algorithm, which personalizes the global \ngmns/\ngr model by merging additional variables using the client's data. \fngm is robust to data heterogeneity, large number of participants, and limited communication bandwidth. 
% we present an approach to learning Neural Graphical Models (NGMs) based on multiple datasets.
% \urszula{To be revised as we add results below.}\
% \textit{Keywords}: Federated Learning, Graphical models, Deep learning, Learning representations\\
%\textit{Software}:{\small\href{https://drive.google.com/drive/folders/1B8T4pRANW_XzK6l1tTdQSIQFwSbQrrOB?usp=sharing}{NGM code link} (provided in Supplementary)}
%{\small\href{https://github.com/Harshs27/neural-graphical-models}{NGM code link}}\urszula{We need to replace it with anonymized code repository link}
\end{abstract}

\section{Introduction and Related Works}
We live in the era of ubiquitous extensive public datasets, and yet a lot of data remains proprietary and hidden behind a firewall. This is particularly true for domains with privacy concerns (e.g., medicine).
% \urszula{More about data distributed in silos, private and proprietary.}
Federated Learning~\cite{mcmahan2017communication,yang2019federated,li2021survey} is a framework for model development that is partially or fully distributed over many clients ensuring data privacy. Our framework, \fngmns, learns a Probabilistic Graphical Model (PGM) utilizing sparse graph recovery techniques by aggregating multiple models trained on private data. 

\textit{FL frameworks.} There are two primary network architectures used for FL. The \textit{centralized} paradigm, where one global model is maintained and the local models are updated periodically, e.g., PFNM~\cite{yurochkin2019bayesian} and FedMA~\cite{wang2020federated}. FedMA uses a Federated Matched Averaging algorithm which performs \textit{neuron matching} to tackle the permutation invariance in the neural network based architectures. Their technique introduces dummy neurons with zero weights while optimizing using the Hungarian matching algorithm. This causes the global model size to increase considerably making it a limitation for their approach. It is important to point out that the FL frameworks are usually developed with keeping 
%some 
specific DL architectures in mind. For instance, it is not straightforward to handle skip connections in FedMA due to the dynamic resizing of neural network layers. The  \textit{decentralized} paradigm performs decoupled learning in a peer-to-peer communication system~\cite{li2021decentralized,sun2022decentralized}.

\fngm can work with both paradigms, but we explain it only within the master-clients architecture.
We focus on \ngmsns/\ngrs as they %are generative models (PGMs) which 
learn the underlying distribution from multimodal data. Their inference capabilities allow efficient calculation of conditional and marginal probabilities which can answer many complex queries. Thus, having efficient FL for PGMs 
%can be favored over their discriminative models equivalent.
allows for more flexibility of use than purely predictive models for a single preselected outcome variable.
%\vspace{-5mm}
\begin{figure*}%[h]
\centering 
\includegraphics[width=120mm]{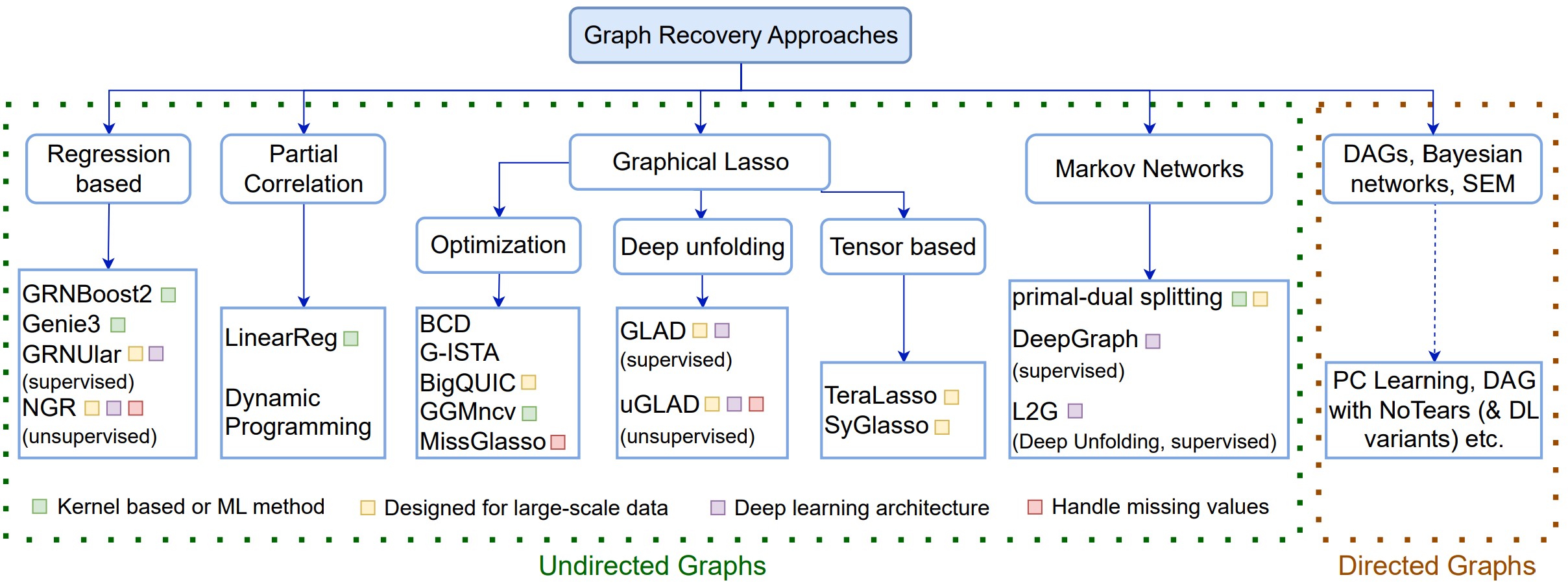}
\caption{\small \textit{Graph Recovery approaches.} Methods that recover graphs are categorized. \fngm typically utilizes undirected graph recovery methods like Neural Graph Revealers(\texttt{NGR})~\cite{shrivastava2023NGR} or the \uglad~\cite{shrivastava2022uglad,shrivastava2022a} algorithm. One can use the methods that retrieve directed graphs by adding a post-processing step of \textit{moralization} which converts them into their undirected counterparts. The algorithms (leaf nodes) listed here are representative of the sub-category and the list is not exhaustive. 
Figure borrowed from \cite{shrivastava2023methods}.
}
\label{fig:graph-recovery-approaches}
\end{figure*}

\textit{Sparse graph recovery.} Given a dataset with $M$ samples and $F$ features, the graph recovery problem aims to recover a sparse graph that captures direct dependencies among the features. Fig.~\ref{fig:graph-recovery-approaches} summarizes the list of approaches for doing sparse graph recovery. %We recommend the going through 
See the survey~\cite{shrivastava2023methods} for more details. 

% More about FL history and state of the art.  What we are doing is different, as we are learning a PGM. Cite neuron matching. Stic

\textit{Choice of Probabilistic Graphical Models.} Graphical models are a powerful tool to analyze data. They can represent the relationship between the features and provide underlying distributions that model functional dependencies between them~\cite{koller2009probabilistic}. %Some of the popular variants are discussed below
% Probabilistic graphical models (PGMs) are quite popular and often used to describe various systems from different domains. Bayesian networks (directed acyclic graphs) and Markov networks (undirected graphs) are able to represent many complex systems due to their generic mathematical formulation~\cite{pearl88,koller2009probabilistic}. These models rely on conditional independence assumptions to make representation of the domain and the probability distribution over it feasible.
% Learning, inference and sampling are operations that make such graphical models useful for domain exploration. Learning, in a broad sense, consists of fitting the distribution function parameters from data. Inference is the procedure of answering queries in the form of marginal distributions or reporting conditional distributions with one or more observed variables. Sampling is the ability to draw samples from the underlying distribution defined by the graphical model. 
We propose using Neural Graphical Models (\ngmsns)~\cite{shrivastava2023NGM} 
%for doing FL. 
within the Federated Learning framework. \ngms accept a feature dependency structure that can be given by an expert or learned from data. It may have the form of a graph with clearly defined semantics (e.g., a Bayesian network or a Markov network) or an adjacency matrix. Based on this dependency structure, \ngms represent the probability function over the domain using a deep neural network. The parameterization of such a network can be learned from data efficiently, with a loss function that jointly optimizes adherence to the given dependency structure and fit to the data. 

Alternatively, we can use Neural Graph Revealers (\ngrsns) which also represent probability function over the domain using a deep neural network.  In contrast to \ngmsns, \ngrs do not require a dependency structure as input to the training algorithm; they recover the structure and learn the network parameterization at the same time with a loss function that jointly optimizes dependency structure sparsity and fit to the data. 

Probability functions represented by \ngms and \ngrs are unrestricted by any of the common restrictions inherent in other PGMs. They also support efficient inference and sampling on multimodal data.
%\urszula{If we drop Appendix A, we need to drop this sentence as well}

% Appendix~\ref{apx:fedngm-related-additional} contains additional background details.
Appendix~\ref{apx:clinical-trials} discusses the use case that motivated this work.

% The rest of this paper is organized as follows:  in Section~\ref{sec:related} we briefly review work most closely related to ours, in Section~\ref{sec:ngm} we summarize Neural Graphical Models and in Section~\ref{sec:fl} we explain representation, learning and inference for Federated Learning with NGMs. We present experiments, both on synthetic and real-life data in Section~\ref{sec:exp}, discuss design considerations and limitations of our framework in Appendix~\ref{apx:design-strategies} and close with conclusions and directions for future work in Section~\ref{sec:conclusions}.

\section{Methods}
%\subsection{Data Setup}

% \urszula{ Also, some of the data assumptions may be novel (not just different distributions, but different feature sets and value sets).}

\textbf{Data Setup}. We have $\mathcal{C}$ clients with proprietary datasets $\{X_1, X_2, ..., X_\mathcal{C}\}$ covering the same domain, where %each $X_c$
%\in\mathbb{R}^{M_c\times F_c}$.   features are not all continuous
each dataset $X_c$ consists of $M_c$ samples, with each sample assigning values to the feature set $F_c$ for the client. The datasets share some, potentially not all, features.  That is, each dataset $X_c$ contains a subset of all features in the domain $F_c \subset F$. Moreover, for some features, value sets overlap and for others they may be completely disjoint. Each client trains a model based on their own data. Note that final client models can differ in their feature sets and value sets. Each user shares model parameters (not data!) and the size of the dataset with the master server. Client datasets remain private. We  assume that all the clients and the master have free access to any  public datasets.

We will explain the procedure using Neural Graphical Models (\ngrsns) first, then cover the changes needed to use it with Neural Graph Revealers (\ngrsns).

\begin{figure*}%[b]
\centering 
\includegraphics[width=120mm]{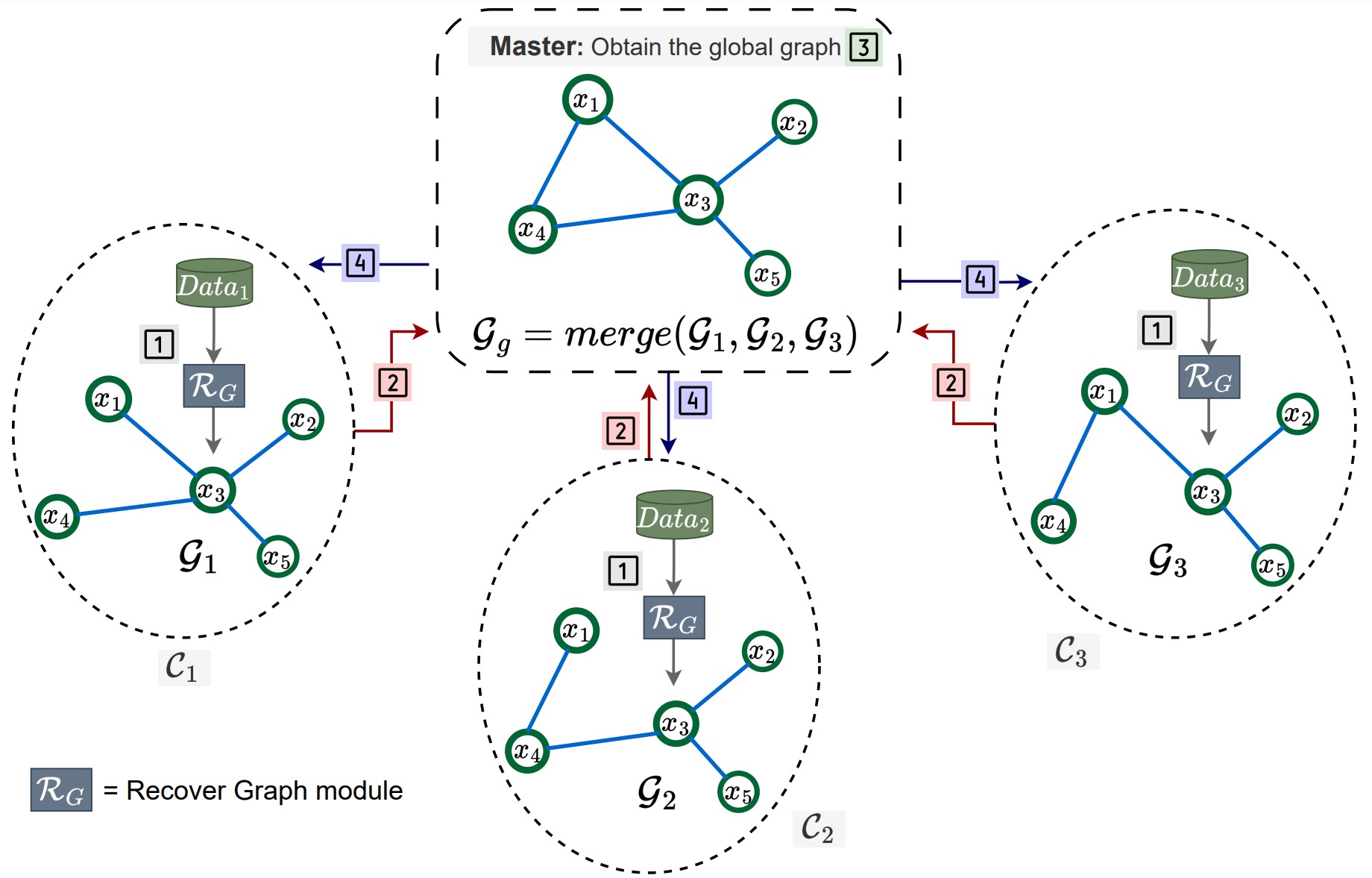}
% \vspace{2mm}
\caption{\small \textit{Obtaining global consensus graph}. In order to set the stage for training individual client \ngmsns, we first obtain the sparse graph that captures the dependency structure among the input features for the entire domain. [1] Each client $\mathcal{C}_c$ runs a graph recovery algorithm (e.g., expert provided, obtained by \ugladns, etc.) on their private data $X_c$,
%\in\mathbb{R}^{M_c\times F_c}$,    they are not all continuous
to obtain a dependency graph with the adjacency matrix $S_c\in\{0,1\}^{F_c\times F_c}$. [2] All the clients send their graphs $S_c$ to the master server. [3] The master server merges all the dependency graphs by only considering the common features across clients  $F_g = \bigcap_{c=1}^C F_c$ and union of all the edges among these common features to obtain the global graph $\mathcal{G}_g$. Please observe the features $x_1, x_2, x_3$ and their connection updates in the master. [4] The master sends the graph $\mathcal{G}_g$ to the clients and then the local and global \ngm models are initialized. One can optionally include public data in this framework.}
\label{fig:fed-graph-init}
% \vspace{-5mm}
\end{figure*}

\subsection{Determining the Global Dependency Graph}

The input to the \ngm model is the data and the corresponding sparse graph that captures feature dependency (\textbf{X},$\mathcal{G}$). We use the sparse graph recovery algorithms like \ugladns, or one of the score-based or constraint-based methods for Bayesian Network recovery or any other algorithm out of an entire suite of algorithms listed in~\cite{shrivastava2023methods}. 

Initially, each client recovers its own sparse graph $\mathcal{G}_c$ based on its own data $X_c$ as described above.  All clients send their graphs to the master.  The master uses the merge function to combine local graphs into the global dependency graph $\mathcal{G}_g$, which is then shared with all clients. Fig.~\ref{fig:fed-graph-init} illustrates this process. 

\textit{Merge function}. This function is used by the master to combine the local dependency graphs obtained from the clients to get the consensus graph $\mathcal{G}_g$. Since we can only gather common knowledge for the features that are available to all the clients, the nodes of the graph $\mathcal{G}_g$ only contain the intersection of features present in the clients' graphs $F_g = \bigcap_{c=1}^C F_c$. We want to make sure that we are not missing out on any potential dependency and for this reason we opt for the conservative choice of including the union of all the edges of the local graphs.

% \urszula{
% The master model will contain the intersection of all features present in the clients' models $F = \bigcap_{c=1}^K F_c$.  For each feature, the set of possible values allowed will be a union of values present in all users' models.  It will be trained based on users' models \urszula{Or samples from users' models?}.  After the master model is trained, each user will receive updated parameters for their model, covering only feature sets and value sets present in their data.  Thus, final models will be personalized for each user.
% }

\begin{figure}%[b]
\centering 
\includegraphics[width=120mm]{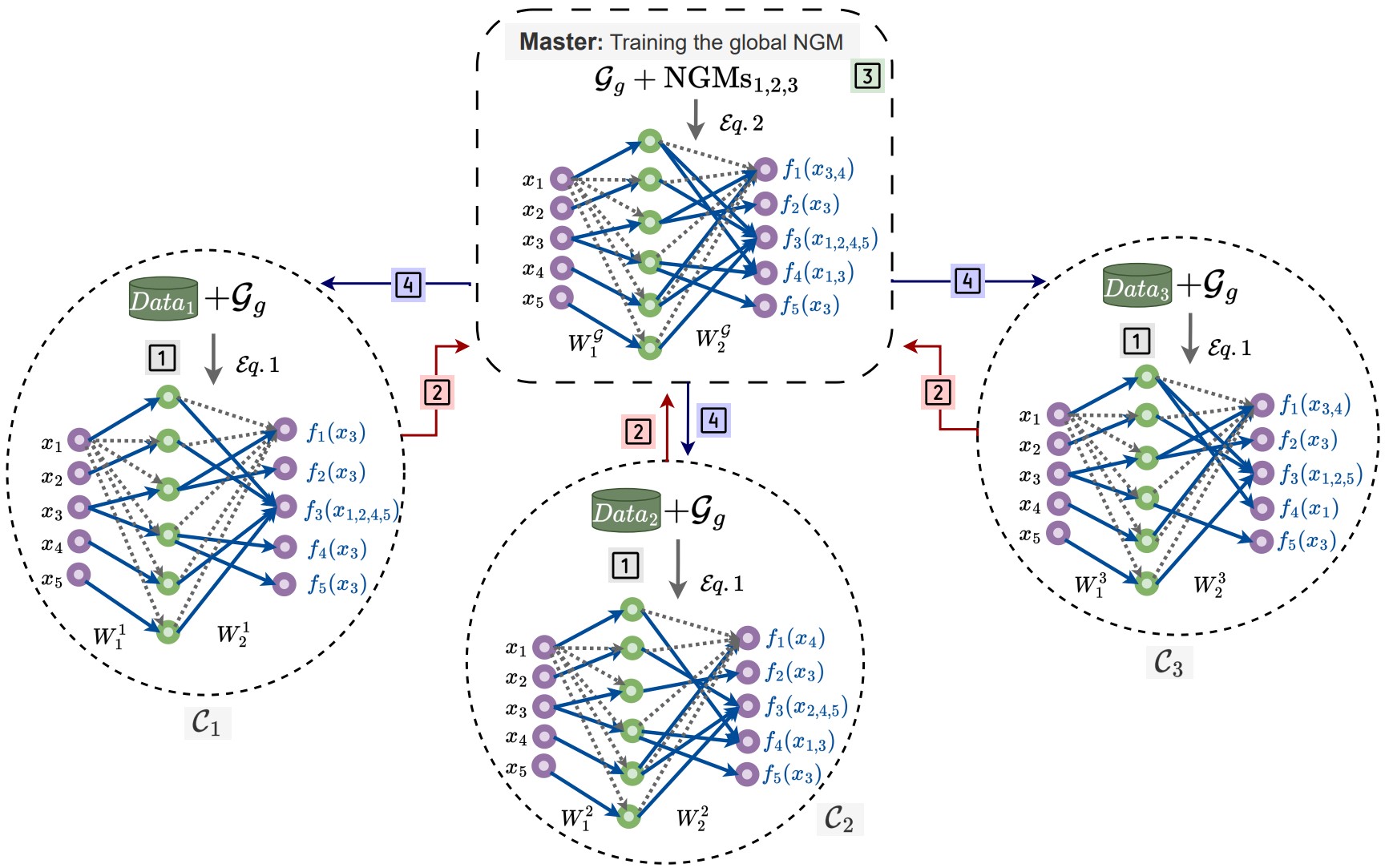}
% \vspace{2mm}
\caption{\small \textit{Training the global \ngmns$_\mathcal{G}$ model}. All the clients and the master have a copy of the global consensus graph $\mathcal{G}_g$. [1] Each client trains the \ngm model based on their data, which is modified to only contain the common features, using the objective in Eq.~\ref{eqn:optimization-local}. [2] All the clients send their trained \ngm models to the master. Note that their data remain private. [3] The master trains the global \ngm model that learns the average of the client \ngms using the objective in Eq.~\ref{eqn:optimization-global}. Optionally, one can use additional public data and samples from client \ngmsns, refer to the inclusion of regression term in Eq.~\ref{eqn:optimization-global-reg}, while training global \ngm for desired results. [4] The global \ngmns$_\mathcal{G}$ model is then transferred to all the clients, where they can run the personalized FL to customize their model using their proprietary data. 
}
\label{fig:fed-training}
% \vspace{-5mm}
\end{figure}

\subsection{Local Training}

Based on the common input global feature graph $\mathcal{G}_g$, the \ngms for the local and global 
%servers 
models are initialized. The architectures of all these \ngms have the same dimensions in terms of the number of hidden units, number of layers and the non-linearity used. 
% We refer the reader to the Appendix~\ref{apx:ngm-review}, which gives insights into these design choices. 

\textit{Local model optimization}. For each client, given its local data \textbf{X}$_c$, the goal is to find the set of parameters $\mathcal{W}$ that minimize the loss expressed as the average distance from individual sample $k$, $X_c^k$ to $f_{\mathcal{W}}(X_c^k)$ while maintaining the dependency structure provided in the input graph \textbf{G} as %. The optimization becomes
\begin{align}\label{eqn:optimization-local}
    \argmin_{\mathcal{W}} \sum_{k=1}^{M_c} \norm{X_c^k - f_{\mathcal{W}}(X_c^k)}^2_2  %\\ \nonumber
    + \lambda \log{\left(\norm{\left(\prod_{i=1}^{L} |W_i|\right) * S_\mathcal{G}^`}_1\right)}
\end{align}
where $S^`$ represents the complement of the master-provided adjacency matrix $S$, which replaces $0$ by $1$ and vice-versa. The $A*B$ is the Hadamard operation. 

\subsection{Training the Global \ngm Model}

%\urszula{We need to rewrite this section}
The master only receives the locally trained \ngm models from its clients and it has no access to their private data. Fig.~\ref{fig:fed-training} outlines a way for the global model to distill knowledge from client models without using any data samples. 
%\urszula{We will also get the number of samples each client model was trained on, so that the master model will properly weigh the contributions of each client model.}

\textit{Global model optimization with samples}. Since each local \ngm has learned a distribution over the same set of features present in the graph $\mathcal{G}_g$, the task of the global model is to learn an average of the distributions represented by the local models. We obtain samples from each of these local models by running the sampling algorithm mentioned in~\cite{shrivastava2023NGM}. We ensure that we get roughly the same number of samples from each of the local models in order to avoid potential biases in learning the global model. The following loss function is optimized
\begin{align}\label{eqn:optimization-global}
    \argmin_{\mathcal{W}^\mathcal{G}}  \sum_{k=1}^{M_c} \norm{X_c^k - f_{\mathcal{W}^\mathcal{G}}(X_c^k)}^2_2 %\\
    + \gamma \log{\left(\norm{\left(\prod_{i=1}^{L} |W_i^\mathcal{G}|\right) * S_\mathcal{G}^`}_1\right)}%\nonumber
\end{align}
The first term adjusts the distribution represented by the global model to be close to the (weighted) average of the client models by optimizing on their samples. The second term is a soft constraint that ensures that the \ngmns's dependency pattern follows that of the consensus graph $\mathcal{G}_g$.

\textit{Additional training (optional)}. We can include public data $X_\mathcal{G}$ as an additional regression term in the objective of  Eq.~\ref{eqn:optimization-global} (represented by $\mathcal{L}_\mathcal{G} (\mathcal{W})$) as 
\begin{align}\label{eqn:optimization-global-reg}
\argmin_{\mathcal{W}}  \mathcal{L}_\mathcal{G} (\mathcal{W})     
 + \sum_{k=1}^{M_\mathcal{G}} \norm{X_\mathcal{G}^k - f_{\mathcal{W}}(X_\mathcal{G}^k)}^2_2 
\end{align}
Note that we can use this additional regression term to address issues like handling \textbf{distribution shifts} in clients' data or doing \textbf{weighted averaging} over the client models. 
% Towards this end, we leverage the sampling ability of the local \ngms and generate the regression data  $X_\mathcal{G}$ for the global model. 
We can balance the data generated as we control the amount of samples from the local \ngm models and thus control the bias while fitting the global \ngm model.

\subsection{Personalized FL for the client models}

\begin{figure}%[b]
\centering 
\includegraphics[width=70mm]{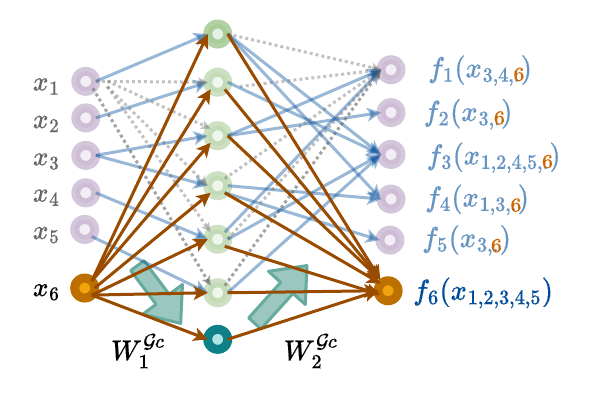}
% \vspace{2mm}
\caption{\small \textit{Personalized FL \textit{Stitching} algorithm}. Each client receives the trained global model \ngmns$_\mathcal{G}$ from the master. The highlighted nodes are the additional nodes introduced in the local model. The nodes in {\color{Bittersweet} orange} are the client specific features, while the {\color{teal} dark green} are the new hidden units introduced to facilitate capturing of dependencies between the common features and the newly added features. Only the new edges ({\color{Bittersweet} orange} arrows and the thick {\color{teal} green} arrows which represent connections to all nodes from the new hidden unit) introduced by the additional nodes are learned from the client's data. One can potentially increase the number of the hidden units for desired results. }
\label{fig:pfl}
\vspace{-5mm}
\end{figure}

Personalized FL aims to customize each client's model to alleviate client drift caused by data heterogeneity. Common approaches for personalizing local models mainly depend on model compression and knowledge distillation~\cite{li2020federated,li2021decentralized,wu2022communication}. In \fngmns, each user receives the global model \ngmns$_\mathcal{G}$, covering only the common feature set across clients. We can  personalize this model when the client data not just have different distributions, but even different feature sets. 

\textit{Stitching algorithm}. Fig.\ref{fig:pfl} explains the process of incorporating client specific features. Specifically, we use the global dependency structure, $\mathcal{G}_g$ augmented by adding client-specific nodes and edges from the client-specific dependency structure $\mathcal{G}_c$ that touch these nodes. We add units for the client-specific features to the input and output layers and additional hidden units to each hidden layer.  We connect all new input nodes to the first hidden layer units and all units in the last hidden layer to the new units in the output layer.  Additionally, we connect all input layer units to the new units in the first hidden layer and all new units in the last hidden layer to all output units.  The connections between hidden layers are added analogously.  We initialize the weights of the new local model as $W_i^{\mathcal{G}c}=W_i^\mathcal{G}\bigcup W_i^E$, where $W_i^E$ represent the weights of the new edges. We retrain the local model, refer to Eq.~\ref{eqn:optimization-local}, using the client's data by freezing the weights $W_i^\mathcal{G}$'s obtained by the global model. The dependency structure is slightly modified $S_{\mathcal{G}c}$ in the graph constraint to account for the additional variables.

%The master model will contain the intersection of all features present in the clients' models $F = \bigcap_{c=1}^C F_c$. It will be trained based on users' models. \urszula{Or samples from users' models?}.

% In the global graph creation step, only the features common to all the clients are retained in the master server. 

%  \harsh{Explain the  here. }

% \harsh{The optimization accounts for maintaining the graph constraints for the added nodes as well. }

% \harsh{----------- START ---------}
% The local variables present in the client's dataset but not part of the master model can be added to the local \ngm as follows:
% \begin{itemize}
%     \item We add a node to the input layer for each private feature
%     \item We add a node to the output layer for each private feature
%     \item We connect each added input node to all nodes in the first hidden layer
%     \item We connect each node in the last hidden layer to all new nodes in the output layer
%     \item We retrain the model using the local dataset
% \end{itemize}

%\urszula{Adding new variables will involve (1) adding nodes to input and output (2) connecting new input and output nodes to all hidden nodes (3) retraining.  Retraining needs to involve some of the old data as the changes in weights between hidden and output will affect them too.}

% \textit{Residual based training}   This will wait for the full paper :-)

% \harsh{----------- END ---------}

\subsection{Using Neural Graph Revealers}

\textbf{Obtaining a Global Variable List.}
In contrast to \ngm case, we do not need to create a global graph before starting local training.  However, to prevent data leakage, the global \ngr model has to involve only variables common to all clients. Therefore, in the first step, the clients send the lists of variables present in their data to the master.  The master compiles the list of common variables $F_g = \bigcap_{c=1}^C F_c$ and sends it back to the clients.

\vspace{2mm}
\textbf{Local Training.}
Initially, each client trains its own \ngr model based on its own data $X_c$ restricted to the global variable list $F_g$.  The architecture of an \ngr model is an MLP that takes in input features and fits a regression to get the same features as the output. We start with a fully connected network.  Some edges are dropped during training.  We view the trained neural network as a glass-box where a path to an output unit (or neuron) from a set of input units means that the output unit is a function of those input units. 
Similarly to the \ngm case, the \ngrs for the local and global 
models are initialized the same way, with the architecture of the have the same dimensions in terms of the number of hidden units, number of layers and the non-linearity used. 

\textit{Local model optimization}. 
The \ngr objective function is designed to jointly discover the feature dependency graph  along with fitting the regression on the input data. The feature dependency graph is defined by network paths.  Our goal is to exclude self-dependencies for all features and induce path sparsity in the neural network.  We observe that the product of the weights of the neural network $S_{nn} = { \prod_{l=1}^L }\abs{W_l}=|W_1|\times |W_2|\times \cdots \times |W_L|$, where $|W|$ computes the absolute value of each element in $W$, gives us path dependencies between the input and the output units. We note that if $S_{nn}[x_i,x_o]=0$ then the output unit $x_o$ does not depend on input unit $x_i$. 

\textit{Graph-constrained path norm.} The \ngr maps NN paths to a predefined graph structure. Consider the matrix $S_{nn}$ that maps the paths from the input units to the output units as described above. Assume we are given a graph with adjacency matrix $S_g\in\{0, 1\}^{D\times D}$. The graph-constrained path norm is defined as $\mathcal{P}_c=\norm{S_{nn}*S_g^c}_1$, where $S_g^c$ is the complement of the adjacency matrix $S_g^c=J_D - S_g$ with $J_D\in\{1\}^{D\times D}$ being an all-ones matrix. The operation $Q*V$ represents the Hadamard operator which does an element-wise matrix multiplication between the same dimension matrices $Q$ \& $V$. This term is used to enforce penalty to fit a particular predefined graph structure, $S_g$.

We use this formulation of MLPs to model the constraints along with finding the set of parameters $\{\mathcal{W, B}\}$ that minimize the regression loss expressed as the Euclidean distance between $X$ and $f_{\mathcal{W, B}}(X)$.
%while maintaining the dependency structure provided in the input graph \textbf{G}. 
Optimization becomes 
\begin{align}\label{eqn:ngr-learning}
    \argmin_{\mathcal{W, B}} \sum_{k=1}^{M} \norm{X^k - f_{\mathcal{W, B}}(X^k)}^2_2 %\\\nonumber
     \quad s.t.~~ \operatorname{sym}(S_{nn}) * S_{\text{diag}} = 0
\end{align}
% \displaystyle 
where, $\operatorname{sym}(S_{nn})=\left(\norm{S_{nn}}_2 + \norm{S_{nn}}_2^T\right)/2$ converts the path norm obtained by the NN weights product, $S_{nn}={\prod_{l=1}^L }\abs{W_l}$, into a symmetric adjacency matrix and
% and $S^c$ represents the complement of the matrix $S$, which essentially replaces $0$ by $1$ and vice-versa. 
$S_{\text{diag}}\in\mathbb{R}^{D\times D}$ represents a matrix of zeroes except the diagonal entries that are $1$. Constraint that disallows self-dependencies is thus included as the constraint term in Eq.~\ref{eqn:ngr-learning}. To satisfy the sparsity constraint, we include an $\ell_1$ norm term $\norm{\operatorname{sym}(S_{nn})}_1$ which will introduce sparsity in the path norms.
Note that this second constraint enforces sparcity of $paths$, not individual $weights$, thus affecting the entire network structure.

We model these constraints as Lagrangian terms which are scaled by a $\log$ function. The $\log$ scaling is done for computational reasons as sometimes the values of the Lagrangian terms can go very low. The constants $\lambda, \gamma$ act as a tradeoff between fitting the regression term and satisfying the corresponding constraints. The optimization formulation to recover a valid graph structure becomes

\begin{align}\label{eqn:optimization-function-ngr}
    \argmin_{\mathcal{W, B}} \sum_{k=1}^{M} \norm{X^k - f_{\mathcal{W, B}}(X^k)}^2_2 + \lambda \norm{\operatorname{sym}(S_{nn}) * S_{\text{diag}}}_1  +\gamma \norm{\operatorname{sym}(S_{nn})}_1%\nonumber
 \end{align}

where we can optionally add $\log$ scaling to the structure constraint terms. Essentially, we start with a fully connected graph and then the Lagrangian terms induce sparsity in the graph.
% This will be helpful in cases where input \textbf{G} is not provided. 
%We note that the optimization and the graph recovered depend on the choices of the penalty constants $\lambda, \gamma$. Since our loss function contains multiple terms, the loss-balancing technique introduced in~\cite{rajbhandari2019antman}, can be utilized to get a good initial value of the constants. Then, while running optimization, based on the regression loss value on a held-out validation data, the values of penalty constants can be appropriately chosen.

%\urszula{We may want to shorten this section}

\begin{figure}%[b]
\centering 
\includegraphics[width=120mm]{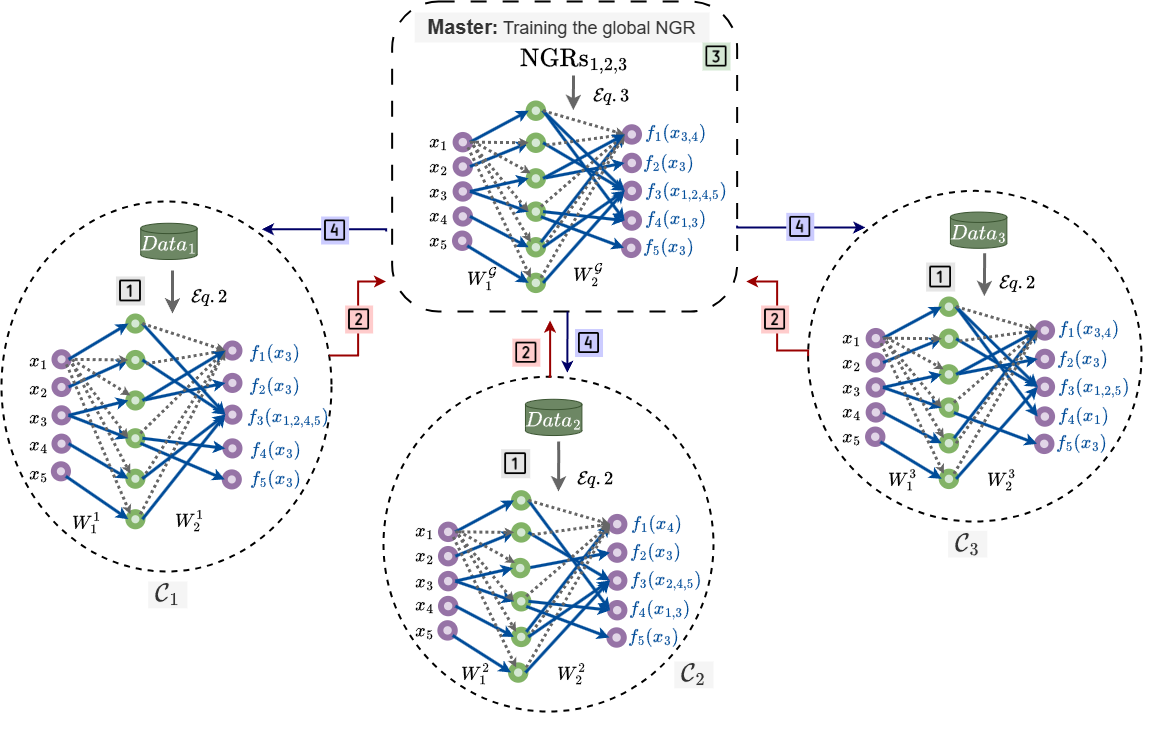}
% \vspace{2mm}
\caption{\small \textit{Training the global \ngr model}. [1] Each client trains the \ngr model based on their data, using the objective in Eq.~\ref{eqn:optimization-function-ngr}. [2] All the clients send their trained \ngr models to the master. Note that their data remain private. [3] The master trains the global \ngr model that learns the average of the client \ngrsns. 
%using the objective in Eq.~\ref{eqn:optimization-global}. 
%Optionally, one can use additional public data and samples from client \ngrsns, refer to the inclusion of regression term in Eq.~\ref{eqn:optimization-global-reg}, while training global \ngr for desired results. 
[4] The global \ngr model is then transferred to all the clients, where they can run the personalized FL to customize their model using their proprietary data. 
}
\label{fig:fedNGR}
% \vspace{-5mm}
\end{figure}

\textbf{Training the Global \ngr Model}
The master only receives the locally trained \ngr models from its clients and it has no access to their private data. 
The master generates a number of samples from each of the client \ngr models.  The number of samples may be proportional to the original size of the datasets local \ngr models were trained on, if available.  The task of the global model is to learn an average of the distributions represented by the local models. 
%\textit{Global model initialization with linear approximation}. We initialize the weights $\mathcal{W}^\mathcal{G}$ with $L$ layers, of the global \ngr model:
%\begin{align}\label{eqn:initialization-global}
%    \argmin_{\mathcal{W}^\mathcal{G}} \norm{\prod_{i=1}^{L} |W_i^\mathcal{G}| - \frac{1}{C}\sum_{c=1}^\mathcal{C} \prod_{i=1}^{L} |W_i^c|}_2^2 %\\
   % + \gamma \log{\left(\norm{\left(\prod_{i=1}^{L} |W_i^\mathcal{G}|\right) * S_\mathcal{G}^`}_1\right)}%\nonumber
%\end{align}
%where $\gamma$ is the Lagrangian penalty constant. The first term adjusts the distribution represented by the global model to be close to the (weighted) average of the client models. The second term is a soft constraint that ensures that the \ngmns's dependency pattern follows that of the consensus graph $\mathcal{G}_g$.
% Alternatively, in the second term, we can enforce a `hard' constraint  by masking. 
%Note that this initialization will assign non-zero weight values to all weights that are non-zero in at least one client's \ngr model.  
%\urszula{Need to check experimentally whether it helps.  Alternatively, we may start with a fully connected \ngr with random weight assignment.}
The global \ngr is trained on client samples in the same way as the client models, following Eq.~\ref{eqn:optimization-function-ngr}.

%\urszula{We can add additional training on public data once we've rewritten section on global training for NGRs}

% \textit{Additional training (optional)}. We can include public data $X_\mathcal{G}$ as an additional regression term in the objective of  Eq.~\ref{eqn:optimization-ngr} (represented by $\mathcal{L}_\mathcal{G} (\mathcal{W})$) as 
% \begin{align}\label{eqn:optimization-global-reg}
% \argmin_{\mathcal{W}}  \mathcal{L}_\mathcal{G} (\mathcal{W})     
%  + \sum_{k=1}^{M_\mathcal{G}} \norm{X_\mathcal{G}^k - f_{\mathcal{W}}(X_\mathcal{G}^k)}^2_2 
% \end{align}
% Note that we can use this additional regression term to address issues like handling \textbf{distribution shifts} in clients' data or doing \textbf{weighted averaging} over the client models. Towards this end, we leverage the sampling ability of the local \ngrs and generate the regression data  $X_\mathcal{G}$ for the global model. We can balance the data generated as we control the amount of samples from the local \ngr models and thus control the bias while fitting the global \ngr model. 

\textbf{Personalized FL for the client models}

Customization of each client's model proceeds similarly to the \ngm case. We add new nodes and edges as before, and retrain the local model, refer to Eq.~\ref{eqn:optimization-function-ngr}, using the client's data by freezing the weights $W_i^\mathcal{G}$'s obtained by the global model, with sparsity contraint applied to new weights only.
%The dependency structure is slightly modified $S_{\mathcal{G}c}$ in the graph constraint to account for the additional variables.

% \subsection{Adding new datasets with overlapping variable sets}
\subsection{Data Privacy}
In our framework, clients' private data are never shared, even with the master.  To further constrain potential data leakage from shared model weights, we can institute a precaution of not sharing either the updated global dependency graph $\mathcal{G}_g$ or the global \ngm master model with any clients unless updates are based on data from no less than $k$ clients.  $k$ will be determined and agreed on with clients depending on the sensitivity of the data in question.  No client should be able to reproduce the other clients' data.

%\harsh{a minimum number of clients so that we do not leak any data. No client should be able to reproduce the other client's data. }

% \section{Experiments}\label{sec:exp}
% \harsh{U - maybe we can add some excerpts from the clinical data here.  A dataset of interest, FLamby: Datasets and Benchmarks for Cross-Silo Federated Learning in Realistic Healthcare Settings~\cite{terrail2022flamby}. Add to experiments section?}
% \urszula{Whatever we put here would be lame, since we didn't do any experiments yet.  Maybe we should just skip this section?}
% % We evaluate NGMs on synthetic and real data. Appendix~\ref{apx:design-strategies} contains some best practices that we developed while working with NGMs. In Appendix~\ref{apx:infant-mortality}, we present an analysis of CDC's \textbf{Infant Mortality} Data~\citep{CDC:InfantLinkedDatasets} using NGMs, which highlights NGMs-generic architecture's ability to model mixed input datatypes.

\section{Experiments}
In the experiments we are using the centralized federated learning paradigm, with the server/master orchestrating steps performed by clients and responsible for aggregating knowledge.  
The same NGM architecture was used for all clients and the global model.  Since we have mixed input data types, real and categorical data, we utilize the \texttt{NGM}-generic architecture~\cite{shrivastava2023NGM}. We used a 2-layer neural view with $H=1000$. The categorical input was converted to its one-hot vector representation and added to the real-valued features which gave us roughly $\sim 500$ features as input.

%\harsh{Any basic info. Centralized FL paradigm etc. Used sampling method to train the global model. The linear method was not working well. Same NGM architecture was used as in the NGM paper. }

\subsection{Infant Mortality data}
The dataset is based on CDC Birth Cohort Linked Birth – Infant Death Data Files \cite{CDC:InfantLinkedDatasets}.  It describes pregnancy and birth variables for all live births in the U.S. together with an indication of an infant's death before the first birthday.  We used the data for 2015 (latest available), which includes information about 3,988,733 live births in the US during 2015 calendar year. The variables include demographic information about parents (age, ethnicity, education), birth order for the current child, pregnancy care and complications, delivery route, gestational age at birth, birth weight, apgar score, newborn care, etc. The dataset is widely used for risk profile analysis for specific causes of death (e.g., SIDS), as well as other undesirable outcomes, such as low birth weight and prematurity.  

\begin{table*}[]
\centering
% \vspace{-2mm}
\caption{\small \textit{Simulated splits on Infant Mortality dataset.} } 
\label{tab:IMsplits}
\resizebox{0.95\textwidth}{!}{
%\small
\begin{tabular}{|c|c|c|}
\hline
Data subsets & Number of patients & Description \\ 
\hline
$\mathcal{C}_1$ & $635,422$ & overrepresentation of mothers with Medicaid and without insurance \\ \hline
$\mathcal{C}_2$ & $645,050$ & overrepresentation of birhts via cesarean section  \\ \hline
$\mathcal{C}_3$ & $645,050$ & overrepresentation of mothers with college degree  \\ \hline
\textit{public} data & $1,925,521$ & distribution matching overall data distribution  \\ \hline
\end{tabular}}
\vspace{-2.5mm}
\end{table*}

\subsection{FL Setup}
We simulate an environment with multiple clients and a global server.  We start by setting aside one half of the dataset, making sure the split is stratified by infant survival, gestational age and the type of isurance used by the mother at birth time.  We refer to this subset as \textit{public} data.  

\begin{figure}
% \vspace{-3.5mm}
\subfigure[]{\label{fig:c1data}\includegraphics[width=0.4\textwidth,height=35mm]{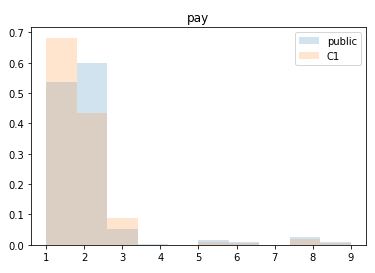}~} %\caption{$\mathcal{C}_1$}}
\subfigure[]{\label{fig:c2data}\includegraphics[width=0.4\textwidth,height=35mm]{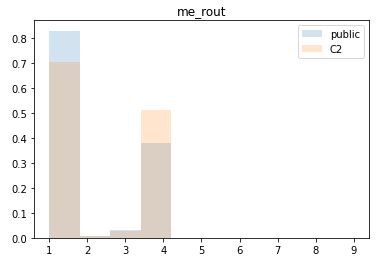}~} %\caption{$\mathcal{C}_2$}}
\subfigure[]{\label{fig:c3data}\includegraphics[width=0.4\textwidth,height=35mm]{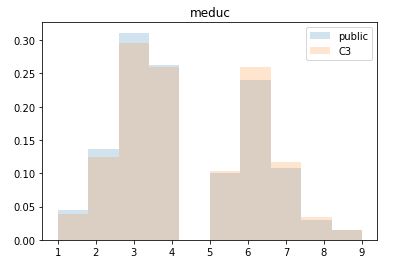}~} %\caption{$\mathcal{C}_3$}}

% \vspace{-6mm}
\caption{\small Infant Mortality dataset was split into public data with distribution matching the entire dataset and three private datasets: $C_1$ with overrepresentation of patients on Medicaid (pay=1) and without insurance (pay=3), $C_2$ with overrepresentation of cesarean section births (me\_rout=4) and $C_3$ with overrepresentation of more highly educated mothers (meduc $\in [5,8]$ indicates at least associate degree).}
\label{fig:IMsplits}
\end{figure}

We select three \textit{client} subsets by oversampling some segments of the population.  Each subset contains over 600,000 patients:
\begin{itemize}
    \item $\mathcal{C}_1$ or private client data 1 contains data with overrepresentation of mothers on Medicaid (variable \textit{pay}, value 1) and mothers with no insurance (variable \textit{pay}, value 3), see~Fig.\ref{fig:c1data}.  We do not have income information in the dataset and the type of insurance is one of the better proxies for affluence.
    \item $\mathcal{C}_2$ or private client data 2 contains data with overrepresentation of deliveries by cesarean section (variable \textit{me\_rout}, value 4), see~Fig.\ref{fig:c2data}.
    \item {$\mathcal{C}_3$} or private client data 3 contains overrepresentation of mothers with college education (variable \textit{meduc}, values 5-8), see~Fig.\ref{fig:c3data}.  
\end{itemize}

% \urszula{TO DO: fix subfigure numbering}

% \harsh{Urszula todo: Explain the various splits defined for the 3 different clients, the ones corresponding to the data you shared. Will be great to have a table with one column as clients $[\mathcal{C}_1, \mathcal{C}_2, \mathcal{C}_3]$ and another column as description and key differentiating factors. }

%\harsh{The global model was trained on the 300K samples (100K from each client).}

Each client trained an \ngm model based on the subset of data \textit{private} to that client.  %\urszula{Harsh, how big was the training data vs test data?}  
The models were passed to the master.  100,000 data samples were generated from each client model, for the total of 300,000 samples.  The global model was trained on these samples.  

We evaluated each client's model performance using 5-fold cross-validation and on the global \textit{public} data.  We compared that performance to the performance of the global model on each of these test sets.

We chose four different variables for assessing predictive accuracy: gestational age (ordinal, in weeks), birthweight (continuous, in grams), survival (binary) and cause of death (multi-valued categorical with 12 values:  10 most common causes of death, less common causes grouped in \textit{other} and \textit{alive}).  

\subsection{Results}
In experiments with models tested on \textit{private} data, we generally observe the global model to match the performance of the private models on real-valued variable prediction and significantly outperform private models on categorical variable prediction.

% \harsh{Urszula todo: Will be great to figure out the best way to showcase the data. Tabular format is not optimal to see the performance comparison. Maybe some kind of plots. }
% \harsh{Urszula todo: Please discuss the results of the tables below}

% \harsh{Under investigation: The dip in the AUPR performance can possibly be explained as it is difficult to get minority class samples for extremely skewed data features, the `survival' feature in this case.}

%\begin{wraptable}[4]{R}{1\textwidth}
\begin{table*}[]
\centering
% \vspace{-2mm}
\caption{\small \textit{Performance on client's private data.} Each client runs their locally trained \ngm model on their proprietary data. Predictive accuracy for gestational age and birthweight using 5fold-CV are shown. The top row shows the client's local model's performance and it is followed by the global-\ngm model's results for that particular client over their entire data. Results are shown for all the other clients. All the client \ngms are learned based on the recovered \uglad graphs.} 
\label{tab:inference_results1}
\resizebox{0.7\textwidth}{!}{
\small
\begin{tabular}{|c|c|c|c|c|}
\hline
Clients & \multicolumn{2}{c|}{Gestational age} & \multicolumn{2}{c|}{Birthweight}  \\ 
& \multicolumn{2}{c|}{(ordinal, weeks)} & \multicolumn{2}{c|}{(continuous, grams)}  \\ \hline
& MAE & RMSE & MAE & RMSE \\ \hline
$\mathcal{C}_1$-\ngm & $1.453 \pm0.017$ & $2.631\pm0.019$ & $373.28\pm0.0$ & $495.24\pm5.75$  \\ \hline
global-\ngm & $1.412 \pm0.0$ & $2.588\pm0.0$ & $377.93\pm0.0$ & $492.26\pm0.0$  \\ \hline\hline
$\mathcal{C}_2$-\ngm & $1.359 \pm0.056$ & $2.392\pm0.066$ & $379.77\pm4.79$ & $497.15\pm5.77$  \\ \hline
global-\ngm & $1.366 \pm0.0$ & $2.403\pm0.0$ & $379.02\pm0.0$ & $499.78\pm0.0$  \\ \hline\hline
$\mathcal{C}_3$-\ngm & $1.427 \pm0.045$ & $2.528\pm0.066$ & $366.47\pm1.90$ & $479.41\pm2.06$  \\ \hline
global-\ngm & $1.37 \pm0.0$ & $2.494\pm0.0$ & $366.08\pm0.0$ & $482.12\pm0.0$   \\ \hline
\end{tabular}}
\label{tab:private-results1}
\vspace{-2.5mm}
%\end{wraptable}
\end{table*}

%\begin{wraptable}[4]{R}{1\textwidth}
\begin{table*}[]
\centering
% \vspace{-2mm}
\caption{\small \textit{Performance on client's private data.} Predictive accuracy for 1-year survival and cause of death are reported. Note: recall set to zero when there are no labels of a given class, and precision set to zero when there are no predictions of a given class. } 
\resizebox{\textwidth}{!}{
\begin{tabular}{|c|c|c|c|c|c|c|}
\hline
Clients & \multicolumn{2}{c|}{Survival} & \multicolumn{4}{c|}{Cause of death} \\ 
& \multicolumn{2}{c|}{(binary)} & \multicolumn{4}{c|}{(multivalued, majority class frequency $0.9948$)} \\ \hline
& \multicolumn{2}{c|}{} & \multicolumn{2}{c|}{micro-averaged}  & \multicolumn{2}{c|}{macro-averaged}  \\ 
& AUC & AUPR & Precision & Recall & Precision & Recall\\ \hline
$\mathcal{C}_1$-\ngm & $0.728\pm0.038$ & $0.256\pm0.048$ & $0.994\pm3.6\text{e-}4$ & $0.994\pm3.6\text{e-}4$ & $0.497\pm1.8\text{e-}4$ & $0.500\pm1.0\text{e-}6$\\ \hline
global-\ngm & $0.797 \pm0.0$ & $0.251 \pm0.0$ & $0.994 \pm0.0$ & $0.994 \pm0.0$ & $0.692 \pm0.0$ & $0.511 \pm0.0$\\ \hline\hline
$\mathcal{C}_2$-\ngm& $0.735\pm0.033$ & $0.242\pm0.013$ & $0.994\pm1.0\text{e-}4$ & $0.995\pm1.0\text{e-}4$ & $0.497\pm5.4\text{e-}5$ & $0.500\pm1.0\text{e-}6$\\ \hline
global-\ngm & $0.799\pm0.0$ & $0.240\pm0.0$ & $0.995\pm0.0$ & $0.995\pm0.0$ & $0.660\pm0.0$ & $0.510\pm0.0$\\ \hline\hline
$\mathcal{C}_3$-\ngm & $0.732\pm0.023$ & $0.293\pm0.027$ & $0.995\pm2.0\text{e-}4$ & $0.995\pm2.0\text{e-}4$ & $0.496\pm1.0\text{e-}4$ & $0.500\pm1.0\text{e-}6$ \\ \hline
global-\ngm & $0.795\pm0.0$ & $0.253\pm0.0$ & $0.995\pm0.0$ & $0.995\pm0.0$ & $0.657\pm0.0$ & $0.511\pm0.0$ \\ \hline
\end{tabular}}
\label{tab:private-results2}
\vspace{-2.5mm}
%\end{wraptable}
\end{table*}

\begin{figure}
% \vspace{-3.5mm}
\subfigure{\label{fig:private_survival_AUC}\includegraphics[width=0.31\textwidth,height=35mm]{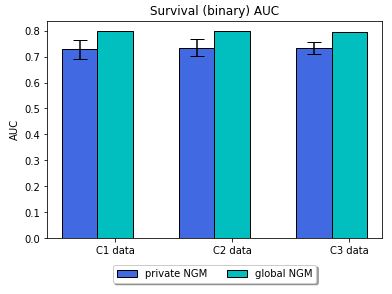}~}
\subfigure{\label{fig:private_cod_precision}\includegraphics[width=0.31\textwidth,height=35mm]{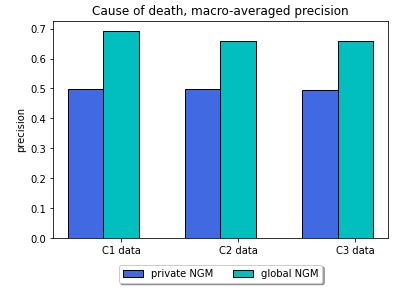}~}
\subfigure{\label{fig:private_cod_recall}\includegraphics[width=0.31\textwidth,height=35mm]{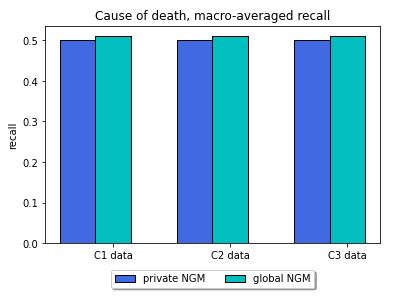}~}

% \vspace{-6mm}
\caption{\small Performance on client's private data.  Predictive accuracy for 1-year survival and cause of death. The same results are presented in Table~\ref{tab:private-results2}}
\label{fig:private}
\end{figure}

In Tables~\ref{tab:private-results1} and \ref{tab:private-results2} we present detailed results on \textit{private} data.  Each time, we compare the performance of private models trained on data coming from the same distribution as the test data to the performance of the global \ngm model.  We use mean absolute error (MAE) and root mean square error (RMSE) as performance metrics for real-valued variables. For the binary variable \textit{survival} we report area under the ROC curve (AUC) and area under the precision recall curve (AUPR).  For the multi valued categorical variable \textit{cause of death}, we report micro- and macro-averaged precision and recall.  We are particularly interested in macro-averaged metrics, as they better capture performance on rare classes. Figure~\ref{fig:private} illustrates some of the results.  Despite the difference in training set size (480,000 for the clients, 300,000 samples for the global \ngm), and potential distribution mismatch, the global \ngm preforms either at par or exceeds the performance of individual client models.

In Table~\ref{tab:public-results1}, we report performance results on \textit{public} data.  In this case, the global \ngm model still has the disadvantage of being trained on a smaller dataset, but the distribution of the test data should more closely match the samples the global \ngm model was trained on than the training data available to clients.  

We observe again a comparable performance between the global \ngm and client models, with the global model typically outperforming slightly at least two of the three client models on real-valued variable prediction and significantly outperforming client models in \textit{survival} AUC and macro-averaged precision and recall for \textit{cause of death}.

%\urszula{We may want to discuss the reasons for superior performance on categorical variables and poor performance on AUPR}

%\begin{wraptable}[4]{R}{1\textwidth}
\begin{table*}[]
\centering
% \vspace{-2mm}
\caption{\small \textit{Performance on public data.}  Models are ran on the public (held-out test) data. Predictive accuracy for gestational age, birthweight, 1-year survival and cause of death are reported. All the client \ngms are learned based on the recovered \uglad graphs. Note: recall set to zero when there are no labels of a given class, and precision set to zero when there are no predictions of a given class.} 
\resizebox{\textwidth}{!}{
\begin{tabular}{|c|c|c|c|c|c|c|c|c|c|c|c|}
\hline
Clients & \multicolumn{2}{c|}{Gestational age} & \multicolumn{2}{c|}{Birthweight} & \multicolumn{2}{c|}{Survival} & \multicolumn{4}{c|}{Cause of death} \\ & \multicolumn{2}{c|}{(ordinal, weeks)} & \multicolumn{2}{c|}{(continuous, grams)} &  \multicolumn{2}{c|}{(binary)} & \multicolumn{4}{c|}{(majority class frequency $0.9948$)} \\ \hline
 &\multicolumn{2}{|c|}{} &\multicolumn{2}{|c|}{} &\multicolumn{2}{|c|}{} & \multicolumn{2}{c|}{micro-averaged}  & \multicolumn{2}{c|}{macro-averaged}  \\ 
& MAE & RMSE & MAE & RMSE & AUC & AUPR & Precision & Recall & Precision & Recall\\ \hline
$\mathcal{C}_1$-\ngm & $1.398$ & $2.532$ & $372.28$ & $489.38$ & $0.711$ & $0.272$ & $0.994$ & $0.994$ & $0.497$ & $0.500$\\ \hline
$\mathcal{C}_2$-\ngm & $1.283$ & $2.389$ & $377.90$ & $491.84$ & $0.757$ & $0.239$ & $0.994$ & $0.995$ & $0.496$ & $0.500$\\ \hline
$\mathcal{C}_3$-\ngm & $1.432$ & $2.542$ & $377.43$ & $493.64$ & $0.775$ & $0.285$ & $0.994$ & $0.995$ & $0.497$ & $0.500$ \\ \hline
global-\ngm & $1.384$ & $2.494$ & $373.29$ & $492.30$ & $0.799$ & $0.241$ & $0.995$ & $0.995$ & $0.674$ & $0.511$\\ \hline
\end{tabular}}
\label{tab:public-results1}
\vspace{-2.5mm}
%\end{wraptable}
\end{table*}

\begin{figure}
% \vspace{-3.5mm}
\subfigure{\label{fig:public_survival_AUC}\includegraphics[width=0.31\textwidth,height=35mm]{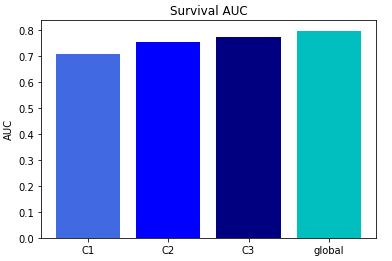}~}
\subfigure{\label{fig:public_brthwgt_RMSE}\includegraphics[width=0.31\textwidth,height=35mm]{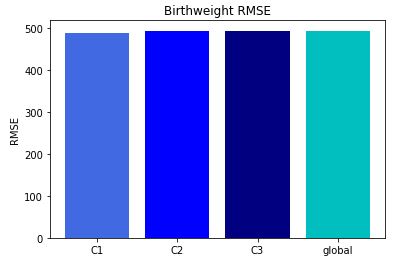}~}
\subfigure{\label{fig:public_combgest_RMSE}\includegraphics[width=0.31\textwidth,height=35mm]{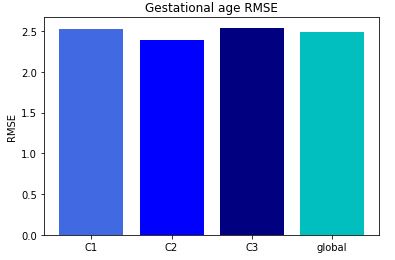}~}

% \vspace{-6mm}
\caption{\small Performance on global test data, with distribution similar to the union of private data.  Predictive accuracy for 1-year survival and root mean squared error for birthweight and gestational age. The same results are presented in Table~\ref{tab:public-results1}}
\label{fig:ggm-expt}
\end{figure}

% \vspace{-3mm}
\section{Conclusions and Future work}\label{sec:conclusions}
% \vspace{-3mm}
This work proposes a Federating Learning framework for \ngmsns, a type of Probabilistic Graphical Model that can handle multimodal data efficiently. PGMs are more flexible than predictive models, 
% with a single outcome variable
as they learn a distribution over all features in a domain and can answer queries about any variable's probability conditional on an assignment of values to any other feature or set of features.  Extending the PGM framework to handle multiple private datasets will allow for knowledge sharing without data sharing, a critical goal in sensitive domains, such as healthcare. \fngms can train a global master model, capturing all domain knowledge and individual clients can train personalized models on their data.\\

\textbf{Applications in video representation \& generation}: We contemplate an interesting application of \fngms that will leverage its privacy preserving federated learning setup to learn video representations and even have an enhanced capability of video generation. Videos can be considered as a sequential collection of frames~\cite{oprea2020review}. \ngrs and \ngms are graph based methods that have a capability to model the feature interactions in a given frame as well as find out similarities as the sequence of frame evolves~\cite{shrivastava2023neural,shrivastava2024methods}. Consider that a \ngr model is learned with each client, such that it captures some aspect like feature connections for video frame understanding. A collection of such models at the master node, can be useful in distilling the properties of the videos present with each of the client in a privacy preserving way. Such a technique can be used to extract common traits for variety of video tasks as prediction of interaction activity, action recognition, video segmentation and object detection~\cite{jiao2021new,bodla2021hierarchical,saini2022recognizing}. \ngrsns-generic architecture's multi-modal handling capacity can be leveraged for modeling continuous multi-dimensional processes~\cite{shrivastava2024video1}, video prior representation~\cite{shrivastava2024video2,shrivastava2024video3} as well as synthesizing diverse video ~\cite{denton2018stochastic,shrivastava2021diverse,shrivastava2021diversethesis} based on the different client's local models. Similar idea of garnering insights using \fngms can be used in conjunction with text mining tools~\cite{fize2017geodict,roche2017valorcarn,antons2020application}, which can play vital roles in extracting information in an efficient and privacy preserving manner.

\bibliography{citations,bibfile}
\bibliographystyle{ECML2023_files/splncs04}
\clearpage
\appendix

\section{Motivating Use Case: Clinical Trials}
\label{apx:clinical-trials}

Clinical trials explore the safety and efficacy of medical interventions: drugs, procedures, devices and treatments.  They are run as randomized controlled experiments with treatment group(s) and control group(s) with carefully screened participants. A detailed evaluation and comparison between groups (often including also subgroups) is performed at the end.  Clinical trials proceed in three phases, moving to the next phase usually requires an FDA (or analogous agency) approval.  Phase 1 evaluates treatment's safety and mechanism at different dosage levels among healthy volunteers.  Phase 2 establishes treatment's benefits and side effects among a relatively small group of people with the disease being treated.  Phase 3 involves a large cohort of people with the disease.  Phase 4 runs after the drug has been approved to assess long-term efficacy and safety.

In the US, ClinicalTrials.gov is a database of privately and publicly funded clinical studies conducted around the world. In addition to intervention trials described above, it includes data on observational trails.  Clinicaltrialsregister.eu is the analogous database maintained in the European Union.

Since 2007, it has been mandated by law that anyone sponsoring a clinical trial in the US register it at ClinicalTrials.gov and report a summary of the results within 1 year after finishing data collection for the trial's primary endpoint or the trial's end, whichever is sooner.  A recent work~\cite{NEJMsa1409364} has found that out of more than 13 thousand trials terminated or completed from January 1, 2008 to August 31, 2012, only 13.4\% reported results within mandated time and 38.3\% reported results at any time. Similar results have been published for European clinical trials~\cite{BMJ.k3218}.  It also reports "extensive evidence of errors, omissions and contradictory data" in the EU registry entries.

Failure to report results may limit the usefulness of any models trained on ClinicalTrials.gov data and their European Union counterpart EU CTR (www.clinicaltrialsregister.eu). In particular, it is possible that withheld data is differently distributed over trial characteristics than reported data.  For example, phase 4 trials are much more likely to report than earlier phase ones and large trials (> 500 participants) more likely to report than smaller ones~\cite{NEJMsa1409364}.

Clinical trial data includes the following information. Most of the fields contain text strings and different studies report information in slightly different formats.
\begin{itemize}[leftmargin=*,nolistsep]
    \item metainformation: sponsor, dates, locations, target duration, type of study (interventional or observational), study phase and status, and description
    \item enrollment criteria: age, gender, disease status, inclusion of healthy volunteers, exclusion criteria
    \item target disease and treatment
    \item trial protocol, arm description
    \item outcome measures, study results
\end{itemize}

Large pharmaceutical companies which sponsor tens and hundreds of clinical trails may have trial data not yet reported or databases with more detailed trial data than officially available.  Leveraging such data through Federated Learning with privacy guarantees would result in more accurate models for everyone.

One of the benefits of pooling all clinical trial data would be to obtain more accurate assessment of clinical trial success rates for each phase and provide insight into features with most impact on that success. Given that our model is a PGM, we can generate predictions of any variable of interest, including overall success of the trial, successful recruitment of volunteers, assess the probability of treatment being effective, etc.  We can also provide insight into dependencies between variables.

%\urszula{--------------END------------}

%\urszula{Stopping here, since it is not clear to me we can define trial success that everyone will agree to.  It may be easier to predict whether the trial will be completed or terminated.  About 10\% of finished studies are terminated and 15\% of studies with results. It may come to the same thing.  Many trials are terminated because they couldn't find enough volunteers or because early results were bad.  In both cases, it is a failure of the trial.}

%\urszula{Sharing model parameters, feature sets and value sets may be too much already.  We may want to generate synthetic data from a known distribution and add it to model training in a known proportion at each user. When done carefully, it should be enough to make it DP.  To verify}
% \urszula{
% Main design decisions:
% \begin{itemize}
%     \item \textbf{Master model} vs. distributed
%     \item \textbf{Differential privacy}
%     \item \textbf{Personalized} vs. generic
%     \item \textbf{PGM} vs. prediction model
% \end{itemize}

% \urszula{
% Issues:
% \begin{itemize}
%     \item We may have edges (paths) that are dependencies only for some users' graphs
%     \item We need to treat values as first-class citizens, which will probably mean binning continuous variables (signal loss)
% \end{itemize}

% Potential use cases:
% \begin{itemize}
%     \item patent novelty
%     \item NER
%     \item \textbf{clinical trials - predicting successful enrollment or successful outcome of the study}
% \end{itemize}
% }

\end{document}